\documentclass[10pt,twocolumn,letterpaper]{article}

\usepackage{wacv}
\usepackage{times}
\usepackage{epsfig}
\usepackage{graphicx}
\usepackage{amsmath}
\usepackage{amssymb}

\def\wacvPaperID{15} 

\wacvfinalcopy 

\ifwacvfinal
\def\assignedStartPage{1} 
\fi

\ifwacvfinal
\usepackage[breaklinks=true,bookmarks=false]{hyperref}
\else
\usepackage[pagebackref=true,breaklinks=true,colorlinks,bookmarks=false]{hyperref}
\fi

\ifwacvfinal
\else
\fi

\begin{document}

\title{Weakly Supervised Multi-Object Tracking and Segmentation}

\author{
Idoia Ruiz$^\ast$, Lorenzo Porzi$^\dagger$, Samuel Rota Bul\`o$^\dagger$, Peter Kontschieder$^\dagger$, Joan Serrat$^\ast$\\
Computer Vision Center, UAB$^\ast$, Facebook$^\dagger$ \\
{\tt\small \{iruiz,joans\}@cvc.uab.es$^\ast$}
}

\maketitle

\begin{abstract}
We introduce the problem of weakly supervised Multi-Object Tracking and Segmentation, \ie joint weakly supervised instance segmentation and multi-object tracking, in which we do not provide any kind of mask annotation. To address it, we design a novel synergistic training strategy by taking advantage of multi-task learning, \ie classification and tracking tasks guide the training of the unsupervised instance segmentation. For that purpose, we extract weak foreground localization information, provided by Grad-CAM heatmaps, to generate a partial ground truth to learn from. Additionally, RGB image level information is employed to refine the mask prediction at the edges of the objects. We evaluate our method on KITTI MOTS, the most representative benchmark for this task, reducing the performance gap on the MOTSP metric between the fully supervised and weakly supervised approach to just 12$\%$ and 12.7 $\%$ for cars and pedestrians, respectively.
\end{abstract}

\section{Introduction}

Computer vision based applications often involve solving many tasks simultaneously. For instance, in a real-life autonomous driving system, tasks regarding perception and scene understanding comprise the problems of detection, tracking, semantic segmentation, etc. In the literature, however, these are usually approached as independent problems. This is the case of multi-object tracking and instance segmentation, which are usually evaluated as disjoint tasks on separate benchmarks.
The problem of Multi-Object Tracking and Segmentation (MOTS) was recently defined in
\cite{Voigtlaender19CVPR_MOTS}. As an extension of the Multi-Object Tracking problem to also comprise instance
segmentation, it consists in detecting, classifying, tracking and predicting pixel-wise masks for the object
instances present along a video sequence.

Due to the lack of suitable datasets, the first two MOTS benchmarks were introduced in \cite{Voigtlaender19CVPR_MOTS} in order to assess their model, which were annotated manually. The annotation procedure involves providing bounding boxes and accurate pixel-level segmentation masks for each object instance of predefined classes, plus an unique identity instance tag, consistent along the video sequence. Moreover, this needs to be done on a significant amount of data to effectively train a MOTS model. This results in a high annotation cost and makes infeasible to perform it manually. This issue can be mitigated by investigating approaches that do not require all this data to solve the MOTS task. In this work, we address this unexplored line of research.

\begin{figure}
\centering
\includegraphics[width=0.45\textwidth]{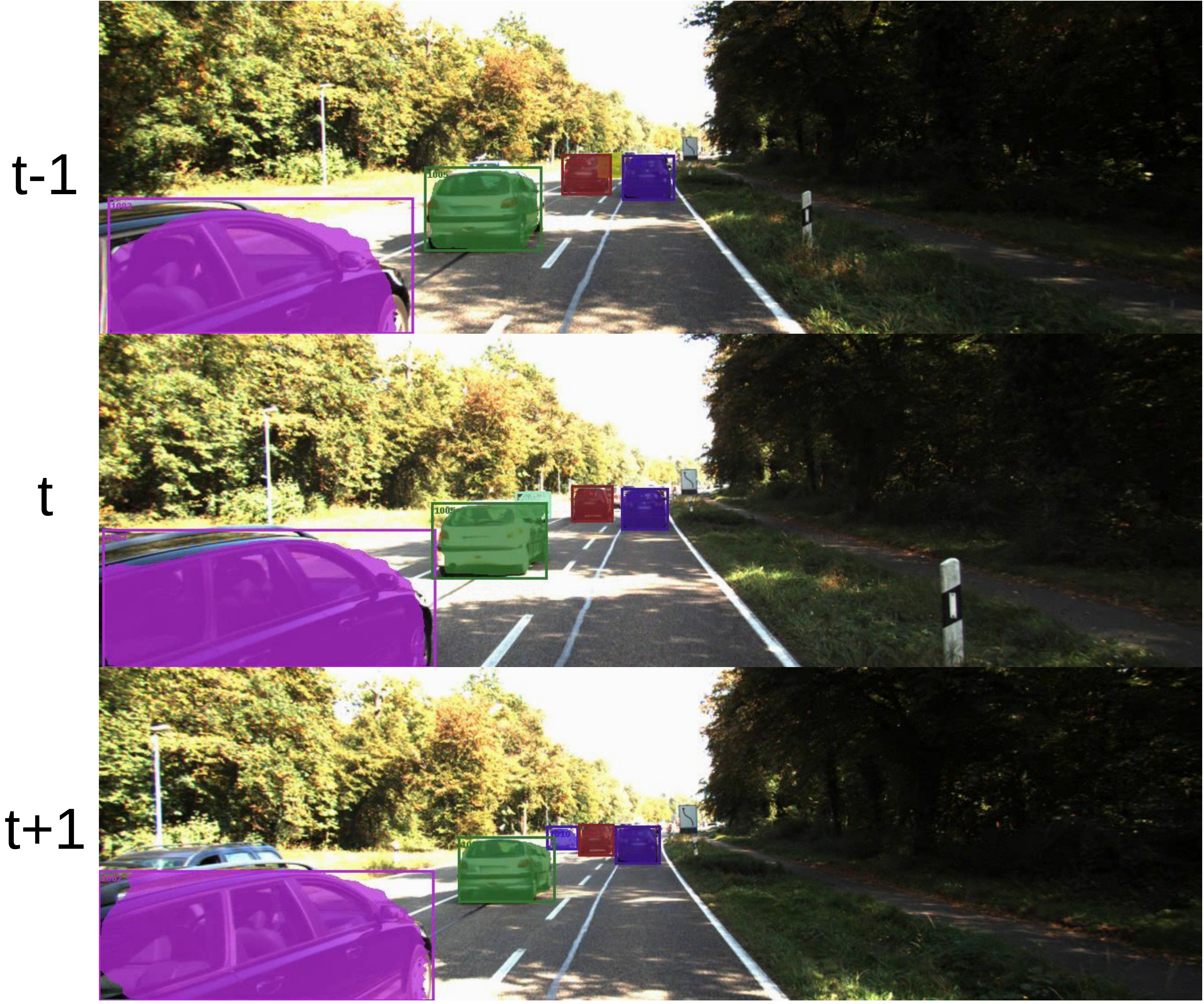}
\caption{Output of our weakly supervised approach on KITTI MOTS. Different colors represent the different identities.}
\label{fig:outputres}
\end{figure}

We define the weakly supervised MOTS problem as the combination of weakly supervised instance segmentation and multi-object tracking. It aims at detecting, classifying, tracking and generating pixel-wise accurate masks, without providing any kind of instance segmentation annotation, the most expensive annotation type of MOTS datasets. We propose an approach that solves this task by using only detection and tracking annotations: bounding boxes along with their corresponding classes and identities.
By taking advantage of multi-task learning, we design a synergistic training scheme where the supervised tasks support the unsupervised one. We are able to solve the instance segmentation task by relying on the learning of the parallel supervised tasks (see Fig. \ref{fig:outputres} for an output example). Specifically, we provide weak supervision from the classification and tracking tasks, along with RGB image level information. The learning of the instance segmentation task solely depends on this novel supervision. The proposed weak supervision consists of three losses that integrate: localization information via activation heatmaps extracted from the classification task, tracking information and RGB image level information, to refine the prediction at the objects boundaries. To the best of our knowledge, this is the first work that solves the MOTS problem under a weakly supervised setting.

Our main contributions are the following:
\begin{itemize}
\item We define the weakly supervised MOTS problem as joint weakly supervised instance segmentation and multi-object tracking.
This is the first work that, to the best of our knowledge, considers this variant of the MOTS problem and solves it not using any kind of instance segmentation annotations.
\item We design a novel training strategy to address weakly supervised MOTS. The different branches of our architecture, MaskR-CNN based, act synergistically to supervise the instance segmentation task, \ie classification and tracking actively help segmentation.
\item We compare our method to the fully supervised baseline on the KITTI MOTS dataset, showing that the drop of performance, on the MOTSP metric is just 12\% and 12.7\% for cars and pedestrians, respectively.
\item Finally, we provide an ablation study about the contribution of the components of our approach.
\end{itemize}

\section{Related Works}

\subsection{Multi-Object Tracking and Segmentation}
The MOTS problem was introduced in \cite{Voigtlaender19CVPR_MOTS}. The solution proposed by the authors consists in a MaskR-CNN based architecture that comprises an additional tracking branch that learns an embedding, later used to match the object instances along the frame sequence. Despite it is a recently introduced topic, there already exist works related to the MOTS problem on a fully-supervised setting. In \cite{hurtado2020mopt}, instead of joining the problems of instance segmentation and tracking, they solve jointly panoptic segmentation and tracking. 
A similar idea to our work, in the sense of using multi-object tracking to help other tasks, is presented in \cite{luiten2020track}. On their approach, MOTSFusion, tracking helps 3D reconstruction and vice-versa. Very recently, a new framework has been proposed in \cite{xu2020Segment} along with a new MOTS dataset, APOLLO MOTS. Differently from the previous works, the instance segmentation task is not solved in a two stage manner from the bounding box predictions. Instead, they use the SpatialEmbedding method, which is bounding box independent and faster. An extension is done in \cite{xupointtrack++}.

There are no previous works addressing weakly supervised settings of the MOTS problem. However, stressing the importance of the need of annotations for MOTS, an automatic annotation procedure for MOTS benchmarks was proposed in \cite{porzi2020learning}, where the authors also presented a similar architecture to \cite{Voigtlaender19CVPR_MOTS}. As the result of their automatic annotation pipeline, they obtain instance segmentation masks and tracking annotations. However, the masks are obtained from a network that is previously trained using instance segmentation masks from a different benchmark, with a domain gap presumably small with respect to the target dataset. Our model instead, is trained with no previous knowledge of how a mask "looks like".

\begin{figure*}[t]
\centering
\includegraphics[width=\textwidth]{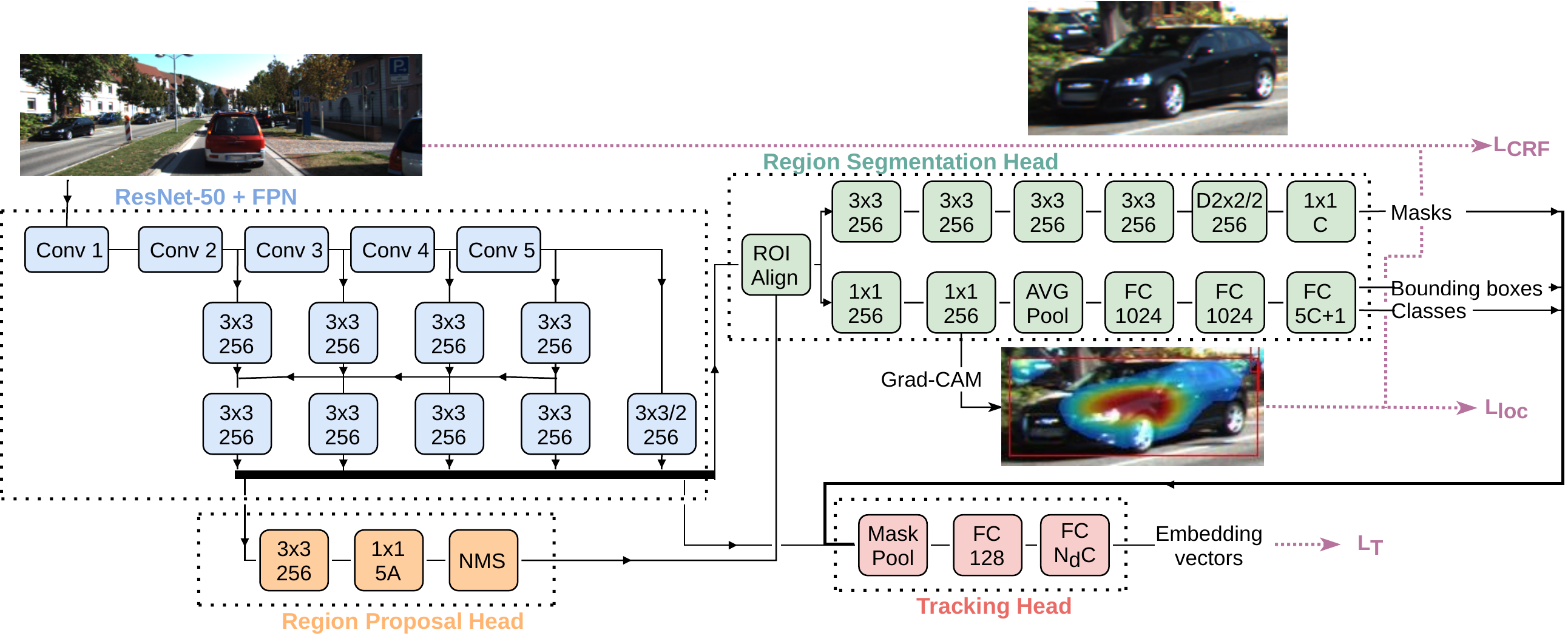}
\caption{Overview of our architecture. We modify MOTSNet \cite{porzi2020learning} by adding 1$\times$1 convolutional layers on the classification and detection branch to extract localization information via Grad-CAM \cite{selvaraju2017grad} heatmaps. We show in purple the losses, $L_{loc}$, $L_{CRF}$ and $L_T$, that supervise the instance segmentation task in the weakly supervised setting.}
\label{fig:architecture}
\end{figure*}

\subsection{Weakly Supervised Segmentation}
The literature in the field of semantic segmentation is extensive and there exist many works that address the weakly supervised setting.
A widely used strategy is to predict an initial weak estimate of the mask, that is then refined by using extra information extracted from the image, \eg using Conditional Random Fields (CRF) as a post-processing step is a common approach to get precise boundaries of the objects.

Some works that follow such strategy are \cite{saleh2016built,kolesnikov2016seed}, which employ a dense CRF \cite{krahenbuhl2011efficient} to improve their mask prediction. In \cite{kolesnikov2016seed}, the authors propose to minimize the KL divergence between the outputs of the network and the outputs of the CRF, while in \cite{saleh2016built}, they smooth their initial mask approximation by using the CRF. They then minimize a loss that computes the difference between the network prediction and the CRF output. Both of them use activations of the network as an initial mask estimation. More recently, \cite{song2019box} employs CRF post-processing to refine initial rectangle-shaped proposals, that are later used to compute the mean filling rates of each class. With their proposed filling rate guided loss, they rank the values of the score map, then selecting the most confident locations for back propagation and ignoring the weak ones.

The mean-field inference of the CRF model \cite{krahenbuhl2011efficient} was later formulated in \cite{crfasrnn_ICCV2015} as a Recurrent Neural Network, which allows to integrate it as a part of a CNN, and train it end-to-end. This formulation is used in the architecture from \cite{arnab2017pixelwise,li2018weakly}. In \cite{arnab2017pixelwise}, it is used to refine the initial semantic segmentation and the final instance segmentation predictions. A weakly supervised panoptic segmentation method is proposed in \cite{li2018weakly}. Two outputs are proposed as the initial masks. If bounding box annotations are available, they use a classical foreground segmentation method. Otherwise, the approximate masks are localization heatmaps from multi-class classification \cite{selvaraju2017grad}, similarly to us. However, their classification network is previously trained and only used to extract the heatmaps. We instead, train all the classification, detection, instance segmentation and tracking tasks simultaneously. Also, we do not have an independent classification network dedicated to extract the heatmaps, it is part of the main architecture. Another advantage of our method is that it extracts the heatmap individually for each ROI proposal, instead of doing it for the whole image.

Differently from the previous methods, the work of \cite{Tang_2018_ECCV}, that considers the problem of training from partial ground truth, integrates the CRF regularizer into the loss function, then avoiding extra CRF inference steps. Their weakly-supervised segmentation loss function is composed by a ground truth plus a regularization term. They propose and evaluate several regularization losses, based on Potts/CRF, normalized cut and KernelCut regularizers.

\subsection{Video Object Segmentation}
Video Object Segmentation (VOS) is a problem related to ours, as it also comprises tracking and segmentation.  In VOS, all the salient objects that appear in the sequence must be segmented and tracked, regardless of their category. Salient objects are those that catch and maintain the gaze of a viewer across the video sequence. Differently, in MOTS, we only track and segment objects that belong to specific classes of interest, therefore needing a classification model. Some recent works in the field of VOS are \cite{cheng2018fast,Sun_2020_CVPR,Zhang_2020_CVPR}.
If we add classification to VOS, then distinguishing object instances, it becomes Video Instance Segmentation (VIS) \cite{vis,Bertasius_2020_CVPR,Lin_2020_CVPR}. The datasets designed to assess this task do not usually present strong multi-object interaction, then lacking hard scenarios with occlusions and objects that disappear and enter again to the scene, as it is characteristic of MOTS benchmarks.

There exist semi and unsupervised approaches of the VOS problem. In the semi-supervised setting the masks of the objects to be tracked are given in the first frame. Only these objects need to be tracked and segmented throughout the rest of the video. The unsupervised approach, however, consists in detecting all the possible objects in the video and track and segment them throughout the whole sequence. The work of \cite{luiten2020unovost} addresses the unsupervised VOS problem with a MaskR-CNN based architecture, trained on COCO. They do the inference for the 80 classes of COCO, using for mask prediction a very low (0.1) confidence threshold, then merging the mask predicted for all the categories, taking the most confident one when there is overlapping. This method was extended to VIS by just adding classification, also provided by Mask R-CNN.


\section{Method}
\label{sec:method}

We build upon the MOTSNet architecture proposed in \cite{porzi2020learning}. 
It is a MaskR-CNN based architecture with an additional Tracking Head. Its backbone is composed by a ResNet-50 followed by a Feature Pyramid Network which extracts features at different resolutions, later fed to a Region Proposal Head (RPH). The features of the bounding box candidates predicted by the RPH enter the Region Segmentation Head, that learns the classification, detection and instance segmentation tasks and the Tracking Head, that learns an embedding.
We add two 1$\times$1 convolutional layers at the classification and detection branch of the Region Segmentation Head, aimed at Grad-CAM \cite{selvaraju2017grad} computation for the ROI proposals, as described in section \ref{sec:weaklymethod}. This is needed to extract activation information, as the original branch does not include any convolutional layer.
The complete architecture is shown in Fig. \ref{fig:architecture}.

First, we describe the general fully supervised setting to finally introduce our weakly supervised approach. To train the model under a fully supervised setting, we employ the loss function defined in \cite{porzi2020learning}, with minor differences in the tracking loss, described below. The loss function $L$ is then defined as

\begin{equation}
\label{eq:loss}
L = L_T + \lambda (L_{RPH} + L_{RSH})\,,
\end{equation}
where $L_T$, $L_{RPH}$ and $L_{RSH}$ denote the Tracking, Region Proposal Head and Region Segmentation Head losses, respectively. We refer the reader to \cite{porzi2019seamless} for a detailed description of the two latter.

\noindent\textbf{Tracking.}
MOTSNet is based on MaskR-CNN but comprises a new Tracking Head (TH) that learns an embedding at training time and predicts class specific embedding vectors for each proposal. The TH first applies the \textit{mask-pooling} \cite{porzi2020learning} operation on the input features, thereby only considering the foreground of the proposal to compute its embedding vector. This embedding is trained by minimizing a hard-triplet loss \cite{hermans2017defense}, so that instances of the same object are pushed together in the embedding space, while instances of different objects are pushed away. The distance in the embedding space is then used at inference time to associate the proposals and build the tracks. We define the distance as the Cosine distance $d(v, w) = \frac{v \cdot w}{\parallel v \parallel \parallel w \parallel}$ between two embedding vectors $v$ and $w$.

Then, the tracking loss $L_T$ is defined as

\begin{multline}
L_T = \frac{1}{|\mathcal{\breve{R}}|} \sum_{\breve{r} \in \mathcal{\breve{R}}} \max \biggl( \max_{\hat{r} \in \mathcal{\breve{R}} | id_{\hat{r}} = id_{\breve{r}} }  d( a_{\hat{r}}, a_{\breve{r}}) - \\ 
\min_{\hat{r} \in \mathcal{\breve{R}} | id_{\hat{r}} \neq id_{\breve{r}}} d( a_{\hat{r},} a_{\breve{r}}) + \alpha, 0 \biggr)\,,
\end{multline}
where $\mathcal{\breve{R}}$ denotes the set of positive matched region proposals in the batch. The positive proposals are those that match a bounding box from the ground truth with an IoU $>0.5$. $a_{\breve{r}}$ and $id_{\breve{r}}$ stand for the corresponding embedding vector and assigned identity from the ground truth track, of the proposal $\breve{r} \in \mathcal{\breve{R}}$. $\alpha$ is the margin parameter of the hard triplet loss.

At inference time, the tracking association is performed as follows. To link positive proposals from consecutive frames, we first discard those whose detection confidence is lower than a threshold. We then compute a similarity function for each pair of objects. We consider the pairs between the current frame objects and the objects present in the previous frames comprised in a temporal window whose length is previously decided.

The similarity function $Sim(\breve{r},\hat{r})$ of two proposals $\breve{r}$ and $\hat{r}$ takes into account the embedding distance and the bounding box overlapping as

\begin{equation}
Sim(\breve{r},\hat{r}) = IoU(b_{\breve{r}}, b_{\hat{r}})  d(  a_{\breve{r}}, a_{\hat{r}})\,,
\label{eq:cost}
\end{equation}
where $b_{\breve{r}}$, $b_{\hat{r}}$ are the predicted bounding boxes associated to $\breve{r}$ and $\hat{r}$,
respectively. From this similarity, we define a cost
\begin{equation}
Cost(\breve{r},\hat{r}) = \left[\max_{\breve{r}, \hat{r} \in \mathcal{\breve{R}} } Sim(\breve{r},\hat{r})\right] - Sim
(\breve{r},\hat{r})\,.
\end{equation}

Finally, the matching is solved by using the Hungarian algorithm.


\subsection{Weakly supervised approach }
\label{sec:weaklymethod}

The loss function that trains the model under a fully supervised setting is defined in Eq. \ref{eq:loss}, where $L_{RSH}$ is

\begin{equation}
L_{RSH} = L_{RSH}^{cls} + L_{RSH}^{bb} + L_{RSH}^{msk}\,,
\end{equation}
$L_{RSH}^{cls}$, $L_{RSH}^{bb}$ and $L_{RSH}^{msk}$ stand for the classification, bounding box regression and mask segmentation losses of the Region Segmentation Head. In the fully supervised case, $L_{RSH}^{msk}$ corresponds to a cross-entropy loss that compares the instance segmentation ground truth to the predicted masks.

In our weakly supervised setting, we do not have any instance segmentation ground truth available. To train the instance segmentation task, we propose a new approach that benefits from the multi-task design of the MaskR-CNN base architecture, \ie it has a common backbone followed by task-specific heads. We exploit this architecture so that the different branches of MOTSNet act in a synergistic manner, guiding the unsupervised task. In particular, we propose a new definition of $L_{RSH}^{msk}$,
\begin{equation}
L_{RSH}^{msk} = L_{loc} + \lambda_{CRF} L_{CRF}\,,
\end{equation}
where $L_{loc}$ and $L_{CRF}$ stand for the Foreground localization and CRF losses, respectively and $\lambda_{CRF}$ is a regularization parameter.

\begin{figure*}
\centering
\includegraphics[width=0.48\textwidth]{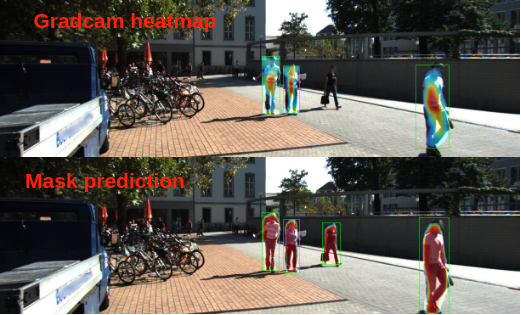}
\hfill
\includegraphics[width=0.48\textwidth]{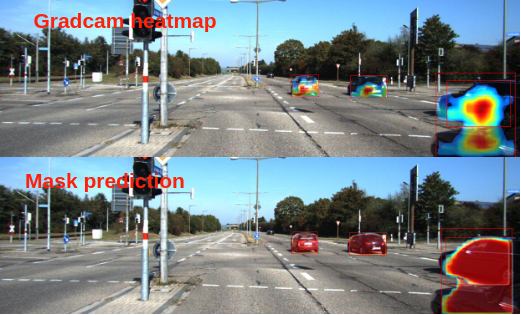}
\caption{Pairs of Grad-CAM heatmaps used as a cue and the corresponding predicted masks.}
\label{fig:gradcammasks}
\end{figure*}

\noindent\textbf{Foreground localization loss $L_{loc}$}.
To provide information to the network about where the foreground is, we use a localization mechanism. In particular, we propose Grad-CAM \cite{selvaraju2017grad}, \ie weak localization heatmaps obtained from the activations and gradients that flow trough the last convolutional layer of a classification network, when it classifies the input as a certain class. Since our architecture naturally comprises a classification branch, we take advantage of that, using the MOTSNet classification branch to compute Grad-CAM heatmaps. As explained in section \ref{sec:method}, we add two 1$\times$1 convolutional layers to the classification and detection branch, before the fully connected layers. The Grad-CAM heatmaps are computed then on the second added convolutional layer by using the implementation variant discussed in section \ref{sec:gradcamanalysis}.

Let $\mathcal{R}$ be the set of bounding boxes from the ground truth. For every bounding box $r \in \mathcal{R}$, we compute the Grad-CAM heatmap $G^r$ corresponding to that ground truth region, for its associated class. We normalize it so that $G^r \in [0,1]^{28\times 28}$. The heatmaps $G^r$ are intended to produce mask pseudo labels to learn from. For a region proposal $\breve{r}$, its corresponding pseudo label $Y^{\breve{r}} \in \{ 0,1,\emptyset\}^{28\times 28}$ is a binary mask generated from the heatmaps, where $\emptyset$ denotes a void pixel that does not contribute to the loss. The assignment of the pseudo label $Y^{\breve{r}}_{ij}$ to the cell $(i,j)$ is defined as
\begin{equation}
Y^{\breve{r}}_{ij} =
\begin{cases}
0 & \forall \, ij \notin \mathcal{P}^{r} \; \; \forall r \in \mathcal{R}\\
1 & \text{if} \; G^{r}_{ij} \geq \mu_{A} \; \; \forall \, ij \in \mathcal{P}^{r} \\
\emptyset & \text{if} \; G^{r}_{ij} < \mu_{A} \; \; \forall \, ij \in \mathcal{P}^{r}\,,
\end{cases}
\end{equation}
where $\mathcal{P}^r$ is the set of pixels that belong to the area defined by the ground truth bounding box $r$. We consider as foreground the pixels of the ground truth bounding boxes whose Grad-CAM value  $G^r$ is above a certain threshold $\mu_{A}$ and background all the pixels outside the bounding boxes. We ignore those pixels that are inside the bounding boxes but below the threshold. Then, the foreground localization loss $L_{loc}$ is a cross entropy loss, defined for a proposal $\breve{r}$ as

\begin{multline}
    L_{loc} ( Y^{\breve{r}}, S^{\breve{r}}) = - \frac{1}{|\mathcal{P}^{\breve{r}}_Y|} \sum_{(i,j) \in \mathcal{P}^{\breve{r}}_Y} Y^{\breve{r}}_{ij} log S^{\breve{r}}_{ij} \\
    - \frac{1}{|\mathcal{P}^{\breve{r}}_Y|} \sum_{(i,j) \in \mathcal{P}^{\breve{r}}_Y} (1 - Y^{\breve{r}}_{ij}) log (1 - S^{\breve{r}}_{ij})\,,
\end{multline}
where $S^{\breve{r}} \in [0,1]^{28\times 28} $ denotes the mask prediction for the proposal $\breve{r}$ for its predicted class, whose entries $S^{\breve{r}}_{ij}$ are the probability of cell $(i,j)$ to belong to the predicted class. $\mathcal{P}^{\breve{r}}_Y \subset \mathcal{P}^{\breve{r}}$ denotes the set of all the non-void pixels in the 28$\times$28 pseudo label mask $Y^{\breve{r}}$, letting $\mathcal{P}^{\breve{r}}$ be the set of all the pixels in $Y^{\breve{r}}$. The loss values of all the positive proposals (those with a bounding box IoU $> 0.5$) are averaged by the number of proposals to compute the loss.

\noindent\textbf{CRF Loss $L_{CRF}$ }.
We use the loss proposed in \cite{Tang_2018_ECCV} to improve the instance segmentation prediction on the object boundaries. This loss integrates CRF regularizers, that can act over a partial input, improving the quality of the predicted mask, then avoiding additional CRF inference steps, as many weakly supervised segmentation methods do \cite{arnab2017pixelwise,li2018weakly,saleh2016built,kolesnikov2016seed}.
The CRF loss $L_{CRF}$ is a regularization loss, result of applying a relaxation of the Potts/CRF model.
According to \cite{Tang_2018_ECCV}, it can be approximated as
\begin{equation}
L_{CRF} (S^{\breve{r}}) = \sum_{k} S^{\breve{r} k'} W(1- S^{\breve{r} k})\,,
\end{equation}
where $W$ represents an \textit{affinity matrix}, \ie the matrix of pairwise discontinuity costs,
$k$ denotes the class and $S^{\breve{r} k} \in [0,1]^{128\times 128}$ is the predicted mask for that class, resized from 28$\times$28 to 128$\times$128 in order to extract quality information from the RGB image. 
Following the implementation of \cite{Tang_2018_ECCV}, we consider a dense Gaussian kernel over RGBXY, then $W$ is a relaxation of DenseCRF \cite{krahenbuhl2013parameter} and the gradient computation becomes standard Bilateral filtering that can be implemented by using fast methods such as \cite{adams2010fast}.
Similarly as with the $L_{loc}$ loss, we average the losses for all the positive proposals.

\noindent\textbf{Tracking loss $L_T$}. As described before, the TH first applies the \textit{mask pooling} operation, \ie the embedding vector predicted by the TH only considers the foreground according to the predicted mask. The tracking loss is then also indirectly supervising the instance segmentation branch.

In summary, the training of the instance segmentation branch is guided by the linear combination of these losses. The algorithm overview is depicted in Fig. \ref{fig:architecture}. The RGB image is used along with the mask prediction to compute $L_{CRF}$, while the ground truth bounding boxes are used to compute Grad-CAM heatmaps that produce pseudo labels to learn from, via a cross-entropy loss applied on the mask prediction. Finally, the TH employs the mask prediction to produce embedding vectors, then indirectly supervising the instance segmentation task. The effect of the combination of the aforementioned losses is shown on Fig. \ref{fig:gradcammasks}, where we show the initial Grad-CAM heatmaps that are used to produce pseudo labels and the final predicted mask by the weakly supervised mask branch.


\subsection{Grad-CAM analysis}
\label{sec:gradcamanalysis}
In the original implementation of \cite{selvaraju2017grad}, the Grad-CAM heatmap $G^c \in \mathbb{R}^{28x28}$ for a certain class $c$ is computed as
\begin{equation}
    G^c = ReLU \biggl( \sum_k \alpha^c_k A^k \biggr)\,,
\end{equation}
where the importance weights $\alpha^c_k $ are defined as the global-average-pooled gradients  $\frac{\partial y^c }{\partial A^k_{ij} } $ over the width and height dimensions $i,j$,
\begin{equation}
    \alpha^c_k = \underbrace{ \frac{1}{Z} \sum_i \sum_j }_{\text{global average pooling}} \frac{\partial y^c }{\partial A^k_{ij} }\,,
\end{equation}
where $y^c$ is the classification score for class $c$ and $A^k$ are the activations of the feature map $k$ of the last convolutional layer in the classification architecture.

\begin{figure}
\centering
\includegraphics[width=0.48\textwidth]{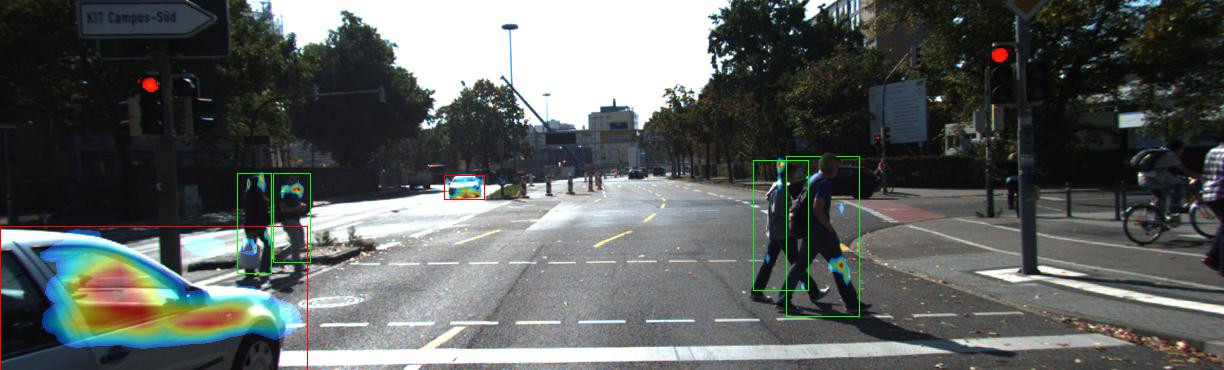}
\includegraphics[width=0.48\textwidth]{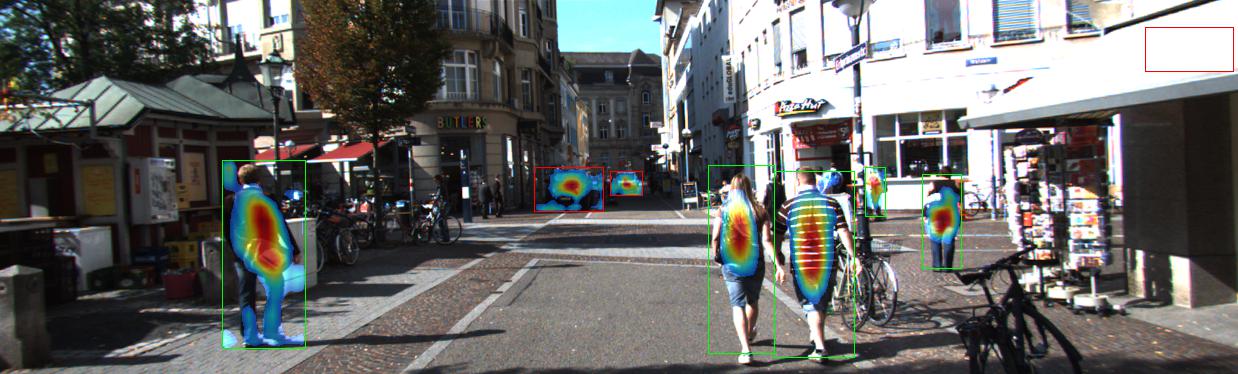}
\caption{Comparison of Grad-CAM heatmaps when using the original Grad-CAM definition (top) and an implementation variant that uses the absolute value of the global-average-pooled gradients (bottom). The activations are color-coded in the heatmap from the lowest (blue) to the highest (red).}
\label{fig:gradcamcomparison}
\end{figure}

We instead, use the absolute value of $\alpha^c_k$ in our implementation, then not needing the ReLU operation. The ReLU is intended to only consider the features that have a positive influence on the class of interest, as negative pixels are likely to belong to other categories, according to the authors.
By using our alternative, we do not discard the weights that are big in magnitude but of negative sign, which in our experiments leaded to better instance segmentation cues.
A comparison of the computed Grad-CAM heatmaps when using both the original implementation and the absolute weights variant is shown in Fig. \ref{fig:gradcamcomparison}. The original Grad-CAM implementation can lead us to incomplete or not so suitable heatmaps to act as an initial approximate of the masks. In our variant, while the highest value is located in the foreground of the object, the high activation areas cover a region of the foreground that can also be useful.

\section{Experiments}
We assess the performance of our method on the most representative MOTS benchmark, KITTI MOTS \cite{Voigtlaender19CVPR_MOTS}. It provides balanced training and validation sets of cars and pedestrians. It is comprised of 21 sequences, extracted from the original KITTI tracking dataset, and a total of 8k frames that contain 11k pedestrian instances and 27k car instances.

\subsection{Metrics}
The MOTS performance is evaluated by the metrics defined in \cite{Voigtlaender19CVPR_MOTS}. The authors proposed an extension of the MOT metrics \cite{bernardin2008evaluating} to assess the instance segmentation performance. Instead of considering the IoU of the predicted bounding boxes with the ground truth, as in the original metrics, they define them in terms of the mask IoU, as follows
\begin{equation}
sMOTSA = \frac{\widetilde{TP} - |FP| - |IDS|}{|M|}
\end{equation}

\begin{equation}
MOTSA = \frac{|TP| - |FP| - |IDS|}{|M|}
\end{equation}

\begin{equation}
MOTSP = \frac{\widetilde{TP}}{|TP|}\,,
\label{eq:motsp}
\end{equation}
where $M$ stands for the set of ground truth masks, $IDS$ is the number of identity switches, $TP$ account for the masks mapped to a ground truth mask with an IoU $> 0.5 $, $ \widetilde{TP} $ is the sum of IoUs between all the predicted and ground truth masks whose IoU is at least 0.5, that is, the sum of the IoUs between the predicted masks counted as TP and their associated ground truth.

MOTSP is a pure segmentation metric; it measures the IoU of the TP predicted masks with the ground truth, which provides a measurement of the segmentation quality alone. MOTSA and sMOTSA also consider the detection and tracking performance, being sMOTSA more restrictive on the instance segmentation contribution. MOTSA only considers the number of predicted masks with an IoU $>0.5$ with the ground truth, while sMOTSA counts the IoU value itself, thus penalizing low IoUs, despite being greater than 0.5.

\subsection{Experimental setup}

\begin{table}
\begin{center}
\begin{tabular}{ll}
\hline\noalign{\smallskip}
Hyperparameter                    & Value             \\
\noalign{\smallskip}
\hline
\noalign{\smallskip}
\multicolumn{2}{l}{\textit{ Training}} \\
Optimizer                         & SGD               \\
Learning rate                     & 0.02              \\
Number of Epochs                  & 150               \\
Total batch size                  & 24                \\
Embedding dimensionality $N_d$    & 8                 \\
Hard triplet loss margin $\alpha$ & 0.2               \\
Loss weight $\lambda_{CRF}$       & $2 \cdot 10^{-7}$ \\
Grad-CAM threshold $\mu_{A}$      & 0.5               \\
\hline
\multicolumn{2}{l}{\textit{ Tracking}} \\
Length of temporal window         & 10                \\
Detection threshold               & 0.9               \\
\hline
\end{tabular}
\end{center}
\caption{Hyperparameters.}
\label{table:params}
\end{table}

To show the effectiveness of our method, our backbone ResNet-50 is just pretrained on ImageNet. Pretraining on other benchmarks significantly boosts the performance of the models, as shown in \cite{porzi2020learning}. However, we are not interested in optimizing a fully supervised baseline but in comparing the proposed weakly supervised approach with respect to the fully supervised baseline under the same pre-training conditions.

On our main experiments, we set the hyperparameters to the values reported in Tab. \ref{table:params}. Training is run on four V100 GPUs with 32GB of memory.


\begin{table*}
\begin{center}
\begin{tabular}{lllllll}
\hline\noalign{\smallskip}
Method & \multicolumn{2}{c}{sMOTSA} & \multicolumn{2}{c}{MOTSA} & \multicolumn{2}{c}{MOTSP} \\
& Car  & Ped  & Car  & Ped  & Car  & Ped  \\
\noalign{\smallskip}
\hline
\noalign{\smallskip}
\multicolumn{7}{l}{\textit{ Fully supervised}} \\
MOTSNet \cite{porzi2020learning} & 69.0 & 45.4 & 78.7 & 61.8 & 88.0 & 76.5 \\
Ours                             & 69.1 & 35.1 & 80.1 & 52.0 & 87.0 & 75.3 \\
\hline
\multicolumn{7}{l}{\textit{ Weakly supervised}} \\
Ours                             & 54.6 & 20.3 & 72.5 & 39.7 & 76.6 & 65.7 \\
\hline
Relative performance drop        & 21.0 & 42.2 & 9.5  & 23.7 & 12.0 & 12.7 \\
\hline
\end{tabular}
\end{center}
\caption{Results of our approach on KITTI MOTS. The ResNet50 backbone is just pretrained on ImageNet for all the models reported.}
\label{table:main}
\end{table*}

\begin{table*}
\begin{center}
\begin{tabular}{lllllll}
\hline\noalign{\smallskip}
Weakly supervised & \multicolumn{2}{c}{sMOTSA} & \multicolumn{2}{c}{MOTSA} & \multicolumn{2}{c}{MOTSP} \\
losses                   & Car  & Ped  & Car  & Ped  & Car  & Ped  \\
\noalign{\smallskip}
\hline
\noalign{\smallskip}
$L_{loc} + L_{CRF} + L_T$ & 49.3 & 13.1 & 67.6 & 32.0 & 75.0 & 64.8 \\
$L_{loc} + L_T$           & 44.3 & 10.2 & 66.9 & 30.7 & 69.6 & 63.5 \\
$L_{loc} + L_{CRF}$      & 55.0 & 11.0 & 73.0 & 31.2 & 76.7 & 62.5 \\
\hline
\end{tabular}
\end{center}
\caption{Results of the ablation study on the weakly supervised approach on KITTI MOTS (run on a previous weaker baseline).}
\label{table:ablation}
\end{table*}

\subsection{Weakly supervised approach}

Since there are no previous works on weakly supervised MOTS, we compare the performance of our weakly supervised approach to the performance of our same model under the fully-supervised setting. To demonstrate that our model can achieve state-of-the-art performance under the supervised setting, we compare it against the current state of the art models under the same training conditions, \ie just pre-training the ResNet-50 backbone on ImageNet. In Tab. \ref{table:main}, on the top section, we compare the performance of our method trained in a fully supervised manner, with the state-of-the-art model \cite{porzi2020learning}. The second section shows the performance of our weakly supervised approach. Our model on both supervised and weakly supervised settings uses the same training hyperparameters (see Tab. \ref{table:params}). When our model is trained on a supervised setting, it achieves slightly superior performance than the state of the art on cars, but is inferior on some metrics for pedestrians. However, MOTSP, defined in Eq. \ref{eq:motsp}, measures the quality of the segmentation masks without taking into account the detection or tracking performance. Our values on this metric, when we train fully supervised, are equivalent to the state of the art on both classes.

\begin{figure}
\centering
\includegraphics[width=0.48\textwidth]{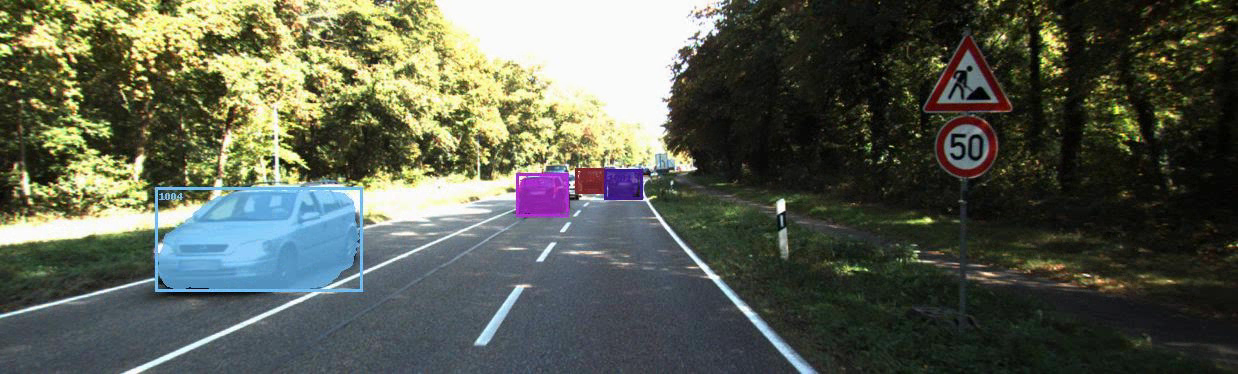}
\includegraphics[width=0.48\textwidth]{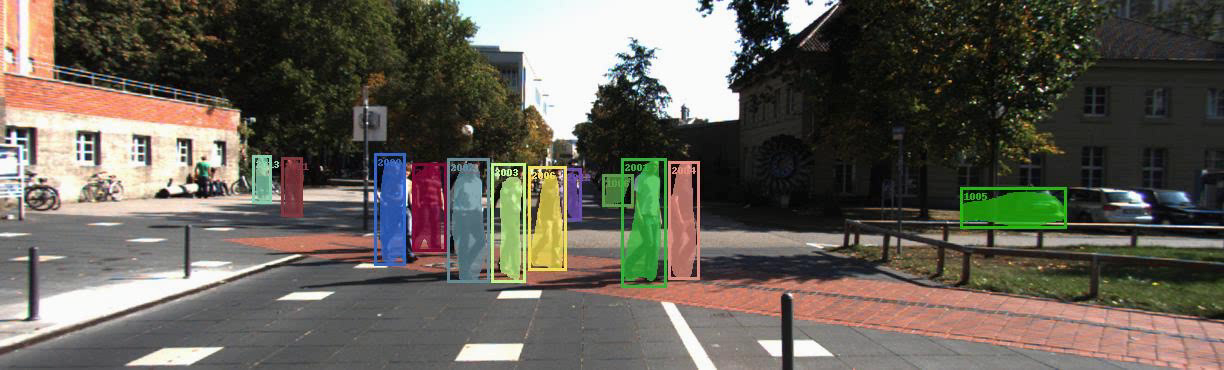}
\caption{Qualitative results on test sequences of KITTI MOTS. Different colors represent the different
identities.}
\label{fig:qualitativeres}
\end{figure}

Finally, the relative drop of performance when training weakly supervised with respect to the supervised case is shown at the bottom line of the table. The performance drop on MOTSP is just a 12.0 $\%$ and 12.7 $\%$ for cars and pedestrians, respectively. This indicates the drop in segmentation quality is not drastic, considering that our model has never been trained with any mask annotation. Regarding MOTSA and sMOTSA, the performance is significantly worse on pedestrians than on cars due to the nature of pedestrians masks. Pedestrians are smaller objects and present more irregular shapes, then retrieving precisely the edges on 128$\times$128 patches is harder. Moreover, Grad-CAM heatmaps can sometimes present high values on the surrounding area of the legs, which leads to incorrect foreground information. Qualitative results are shown on Fig. \ref{fig:qualitativeres}.

\subsection{Weakly supervised approach ablation study}

In order to assess the contribution of our proposed losses to the instance segmentation supervision, we conduct an ablation study in which we test the overall performance when removing the supervision of each loss individually. In the case of the $L_{T}$ loss, we still train the Tracking Head and consider the predicted foreground of the ROIs to compute the tracking embedding vectors, but we do not propagate the gradients to the instance segmentation branch. Thus, we still train the tracking task but it does not affect to the instance segmentation supervision.

On Tab. \ref{table:ablation}, we report the performance of our approach when training with the three losses on the first row. The ablation study was performed in a weaker baseline than our main results from Tab. \ref{table:main}. The second and third row correspond to the experiments, trained with the same hyperparameters, when removing the supervision of $L_{CRF}$ and $L_{T}$ losses, respectively. The $L_{CRF}$ loss clearly helps the supervision, as all the metrics suffer a performance drop when it is not applied. The tracking loss $L_{T}$, however, does help on pedestrians but not on cars. Then, the contribution of the \textit{mask-pooling} layer as a form of supervision on the weakly supervised case is not always positive.


\section{Conclusions}

We have introduced the problem of weakly supervised MOTS, \ie the joint problem of weakly supervised instance segmentation and tracking.
We have contributed a novel approach that solves it by taking advantage of the multitask problem we address. Our architecture is trained in a synergistic manner so that the supervised tasks support the learning of the unsupervised one. In particular, we extract Grad-CAM heatmaps from the classification head, which encode foreground localization information and provide a partial foreground cue to learn from, together with RGB image level information that is employed to refine the prediction at the edges of the objects. We have evaluated our method on KITTI MOTS, the most representative MOTS benchmark, and shown that the drop of performance between the fully supervised and weakly supervised approaches on MOTSP is just a 12 and 12.7 $\%$ for cars and pedestrians, respectively. Finally, we have provided an analysis of the components of our proposed method, assessing their individual contribution.

\section*{Acknowledgements}
Idoia Ruiz and Joan Serrat acknowledge the financial support received for this work from the Spanish TIN2017-88709-R (MINECO/AEI/FEDER, UE) project. Idoia and Joan also acknowledge the support of the Generalitat de Catalunya CERCA Program as well as its ACCIO agency to CVC's general activities.

{\small
\bibliographystyle{ieee_fullname}
\bibliography{egbib}

\begin{thebibliography}{10}\itemsep=-1pt

\bibitem{adams2010fast}
Andrew Adams, Jongmin Baek, and Myers~Abraham Davis.
\newblock Fast high-dimensional filtering using the permutohedral lattice.
\newblock In {\em Computer Graphics Forum}, volume~29, pages 753--762. Wiley
  Online Library, 2010.

\bibitem{arnab2017pixelwise}
Anurag Arnab and Philip~HS Torr.
\newblock Pixelwise instance segmentation with a dynamically instantiated
  network.
\newblock In {\em Proceedings of the IEEE/CVF Conference on Computer Vision and
  Pattern Recognition (CVPR)}, pages 441--450, 2017.

\bibitem{bernardin2008evaluating}
Keni Bernardin and Rainer Stiefelhagen.
\newblock Evaluating multiple object tracking performance: the clear mot
  metrics.
\newblock {\em EURASIP Journal on Image and Video Processing}, 2008:1--10,
  2008.

\bibitem{Bertasius_2020_CVPR}
Gedas Bertasius and Lorenzo Torresani.
\newblock Classifying, segmenting, and tracking object instances in video with
  mask propagation.
\newblock In {\em Proceedings of the IEEE/CVF Conference on Computer Vision and
  Pattern Recognition (CVPR)}, June 2020.

\bibitem{cheng2018fast}
Jingchun Cheng, Yi-Hsuan Tsai, Wei-Chih Hung, Shengjin Wang, and Ming-Hsuan
  Yang.
\newblock Fast and accurate online video object segmentation via tracking
  parts.
\newblock In {\em Proceedings of the IEEE/CVF Conference on Computer Vision and
  Pattern Recognition (CVPR)}, pages 7415--7424, 2018.

\bibitem{hermans2017defense}
Alexander Hermans, Lucas Beyer, and Bastian Leibe.
\newblock In defense of the triplet loss for person re-identification.
\newblock {\em arXiv preprint arXiv:1703.07737}, 2017.

\bibitem{hurtado2020mopt}
Juana~Valeria Hurtado, Rohit Mohan, Wolfram Burgard, and Abhinav Valada.
\newblock Mopt: Multi-object panoptic tracking.
\newblock {\em The IEEE Conference on Computer Vision and Pattern Recognition
  (CVPR) Workshop on Scalability in Autonomous Driving}, 2020.

\bibitem{kolesnikov2016seed}
Alexander Kolesnikov and Christoph~H Lampert.
\newblock Seed, expand and constrain: Three principles for weakly-supervised
  image segmentation.
\newblock In {\em Proceedings of the European Conference on Computer Vision
  (ECCV)}, pages 695--711. Springer, 2016.

\bibitem{krahenbuhl2011efficient}
Philipp Kr{\"a}henb{\"u}hl and Vladlen Koltun.
\newblock Efficient inference in fully connected crfs with gaussian edge
  potentials.
\newblock In {\em Advances in Neural Information Processing Systems}, pages
  109--117, 2011.

\bibitem{krahenbuhl2013parameter}
Philipp Kr{\"a}henb{\"u}hl and Vladlen Koltun.
\newblock Parameter learning and convergent inference for dense random fields.
\newblock In {\em Proceedings of the International Conference on Machine
  Learning}, pages 513--521, 2013.

\bibitem{li2018weakly}
Qizhu Li, Anurag Arnab, and Philip~HS Torr.
\newblock Weakly-and semi-supervised panoptic segmentation.
\newblock In {\em Proceedings of the European Conference on Computer Vision
  (ECCV)}, pages 102--118, 2018.

\bibitem{Lin_2020_CVPR}
Chung-Ching Lin, Ying Hung, Rogerio Feris, and Linglin He.
\newblock Video instance segmentation tracking with a modified vae
  architecture.
\newblock In {\em Proceedings of the IEEE/CVF Conference on Computer Vision and
  Pattern Recognition (CVPR)}, June 2020.

\bibitem{luiten2020track}
Jonathon Luiten, Tobias Fischer, and Bastian Leibe.
\newblock Track to reconstruct and reconstruct to track.
\newblock {\em IEEE Robotics and Automation Letters}, 5(2):1803--1810, 2020.

\bibitem{luiten2020unovost}
Jonathon Luiten, Idil~Esen Zulfikar, and Bastian Leibe.
\newblock Unovost: Unsupervised offline video object segmentation and tracking.
\newblock In {\em The IEEE Winter Conference on Applications of Computer
  Vision}, pages 2000--2009, 2020.

\bibitem{porzi2019seamless}
Lorenzo Porzi, Samuel~Rota Bulo, Aleksander Colovic, and Peter Kontschieder.
\newblock Seamless scene segmentation.
\newblock {\em CoRR}, 2019.

\bibitem{porzi2020learning}
Lorenzo Porzi, Markus Hofinger, Idoia Ruiz, Joan Serrat, Samuel~Rota Bulo, and
  Peter Kontschieder.
\newblock Learning multi-object tracking and segmentation from automatic
  annotations.
\newblock In {\em Proceedings of the IEEE/CVF Conference on Computer Vision and
  Pattern Recognition (CVPR)}, pages 6846--6855, 2020.

\bibitem{saleh2016built}
Fatemehsadat Saleh, Mohammad~Sadegh Aliakbarian, Mathieu Salzmann, Lars
  Petersson, Stephen Gould, and Jose~M Alvarez.
\newblock Built-in foreground/background prior for weakly-supervised semantic
  segmentation.
\newblock In {\em Proceedings of the European Conference on Computer Vision
  (ECCV)}, pages 413--432. Springer, 2016.

\bibitem{selvaraju2017grad}
Ramprasaath~R Selvaraju, Michael Cogswell, Abhishek Das, Ramakrishna Vedantam,
  Devi Parikh, and Dhruv Batra.
\newblock Grad-cam: Visual explanations from deep networks via gradient-based
  localization.
\newblock In {\em Proceedings of the IEEE International Conference on Computer
  Vision (ICCV)}, pages 618--626, 2017.

\bibitem{song2019box}
Chunfeng Song, Yan Huang, Wanli Ouyang, and Liang Wang.
\newblock Box-driven class-wise region masking and filling rate guided loss for
  weakly supervised semantic segmentation.
\newblock In {\em Proceedings of the IEEE/CVF Conference on Computer Vision and
  Pattern Recognition (CVPR)}, pages 3136--3145, 2019.

\bibitem{Sun_2020_CVPR}
Mingjie Sun, Jimin Xiao, Eng~Gee Lim, Bingfeng Zhang, and Yao Zhao.
\newblock Fast template matching and update for video object tracking and
  segmentation.
\newblock In {\em Proceedings of the IEEE/CVF Conference on Computer Vision and
  Pattern Recognition (CVPR)}, June 2020.

\bibitem{Tang_2018_ECCV}
Meng Tang, Federico Perazzi, Abdelaziz Djelouah, Ismail Ben~Ayed, Christopher
  Schroers, and Yuri Boykov.
\newblock On regularized losses for weakly-supervised cnn segmentation.
\newblock In {\em Proceedings of the European Conference on Computer Vision
  (ECCV)}, September 2018.

\bibitem{Voigtlaender19CVPR_MOTS}
Paul Voigtlaender, Michael Krause, Aljosa Osep, Jonathon Luiten, Berin
  Balachandar~Gnana Sekar, Andreas Geiger, and Bastian Leibe.
\newblock Mots: Multi-object tracking and segmentation.
\newblock In {\em Proceedings of the IEEE/CVF Conference on Computer Vision and
  Pattern Recognition (CVPR)}, June 2019.

\bibitem{xu2020Segment}
Zhenbo Xu, Wei Zhang, Xiao Tan, Wei Yang, Huan Huang, Shilei Wen, Errui Ding,
  and Liusheng Huang.
\newblock Segment as points for efficient online multi-object tracking and
  segmentation.
\newblock In {\em Proceedings of the European Conference on Computer Vision
  (ECCV)}, 2020.

\bibitem{xupointtrack++}
Zhenbo Xu, Wei Zhang, Xiao Tan, Wei Yang, Xiangbo Su, Yuchen Yuan, Hongwu
  Zhang, Shilei Wen, Errui Ding, and Liusheng Huang.
\newblock Pointtrack++ for effective online multi-object tracking and
  segmentation.
\newblock {\em The IEEE Conference on Computer Vision and Pattern Recognition
  (CVPR) Workshop on Benchmarking Multi-Target Tracking: Multi-Object Tracking
  and Segmentation}, 2020.

\bibitem{vis}
Linjie Yang, Yuchen Fan, and Ning Xu.
\newblock Video instance segmentation.
\newblock In {\em Proceedings of the IEEE International Conference on Computer
  Vision (ICCV)}, pages 5188--5197, 2019.

\bibitem{Zhang_2020_CVPR}
Yizhuo Zhang, Zhirong Wu, Houwen Peng, and Stephen Lin.
\newblock A transductive approach for video object segmentation.
\newblock In {\em Proceedings of the IEEE/CVF Conference on Computer Vision and
  Pattern Recognition (CVPR)}, June 2020.

\bibitem{crfasrnn_ICCV2015}
Shuai Zheng, Sadeep Jayasumana, Bernardino Romera-Paredes, Vibhav Vineet,
  Zhizhong Su, Dalong Du, Chang Huang, and Philip H.~S. Torr.
\newblock Conditional random fields as recurrent neural networks.
\newblock In {\em Proceedings of the IEEE International Conference on Computer
  Vision (ICCV)}, 2015.

\end{thebibliography}
}

\end{document}